# TACIT KNOW-HOW MADE VISIBLE: USING CONVERSATIONAL AI FOR BUSINESS-PROCESS DOCUMENTATION

*Unnikrishnan Radhakrishnan*

*Department of Business Development and Technology, Aarhus University, Herning, Denmark*
unnik@btech.au.dk

**Abstract**

Small and medium-sized enterprises (SMEs) still depend heavily on tacit, experience-based know-how that rarely makes its way into formal documentation. This paper introduces a large-language-model (LLM)-driven conversational assistant that captures such knowledge on the shop floor and converts it incrementally and interactively into standards-compliant Business Process Model and Notation (BPMN) 2.0 diagrams. Powered by Gemini 2.5 Pro and delivered through a lightweight Gradio front-end with client-side bpmn-js visualisation, the assistant conducts an interview-style dialogue: it elicits process details, asks clarifying questions, and renders live diagrams that users can refine in real time. A proof-of-concept evaluation in an equipment-maintenance scenario shows that the chatbot produced an accurate "as-is" model, highlighted bottlenecks, and generated an improved "to-be" variant, all within a 12-minute session, while keeping API costs within an SME-friendly budget. The study analyses latency sources, model-selection trade-offs, and the challenges of enforcing strict XML schemas, then outlines a roadmap toward agentic, multimodal, and on-premise deployments. The results demonstrate that conversational LLMs can lower the skill and cost barriers to rigorous process documentation, helping SMEs preserve institutional knowledge, enhance operational transparency, and accelerate continuous-improvement efforts.

## 1 Introduction

Process documentation is vital for preserving organisational know-how and ensuring consistent operations, yet many manufacturing-oriented SMEs still rely on tacit knowledge-the unwritten, experience-based expertise embedded in employees' minds—which is hard to articulate and vanishes when seasoned staff depart. The resulting loss of critical know-how threatens operational continuity and product quality, fuelling demand for tools that can swiftly capture and formalise these "head-held" procedures. Recent advances in artificial intelligence, especially large-language-model (LLM) chatbots, now offer a practical avenue: by conversing with domain experts, such agents can elicit process details and transform them into formal Business Process Model and Notation (BPMN) diagrams, lowering the skill barrier for SMEs. This paper therefore examines how LLM-driven chatbots can bridge the documentation gap, beginning with the nature of tacit knowledge and the fundamentals of BPM, and concluding with an evaluation of their effectiveness in real-world SME settings.

1.1 *Tacit knowledge in businesses*: Small and medium-sized enterprises (SMEs) rely on agile, experience-based workflows, yet much of their critical know-how remains undocumented and tacit. Tacit knowledge—the unwritten, experience-based insights people carry in their heads—is notoriously difficult to articulate, as Polanyi famously noted [1]. When veteran staff retire or switch jobs, this undocumented expertise departs with them, risking operational disruption, quality drift, and lengthy onboarding for replacements. Capturing and formalising such "head-held" procedures before they disappear is therefore a strategic imperative for many SMEs.

1.2 *Business Process Modelling:* Business Process Management (BPM) offers a structured approach to mapping, analysing and continuously improving how work is done [2]. By representing tasks, decision points and hand-offs explicitly, BPM helps organisations visualise their operations, identify waste, and embed best practices in daily routines. Research and practice have shown that well-defined process models support consistent execution, training and compliance [3]. Yet SMEs often struggle to adopt BPM: modelling tools require specialised skills, consultants are expensive, and the return on investment may seem uncertain in resource-constrained settings.

1.3 Business Process Model and Notation (BPMN): BPMN 2.0, maintained by the Object Management Group, has become the de facto standard for documenting processes in a form that is both human-readable and machine-interpretable [4]. Its graphical language of events, activities and gateways allow business stakeholders and technical staff to collaborate around the same diagram, and its XML serialisation makes models portable across tools and automation platforms.

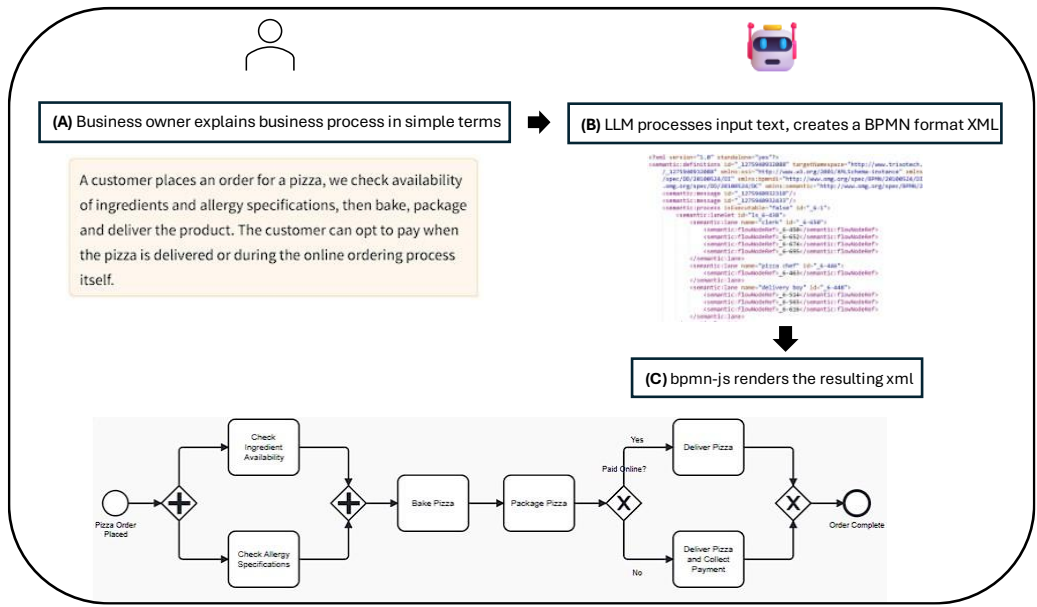

*Figure 1 The core data flow: (A) The user describes a process in natural language. (B) The LLM interprets the text and generates BPMN 2.0 XML. (C) The bpmn-js library renders the XML into a visual diagram.*

The core elements of BPMN include:

- **Flow Objects:** These are the main graphical elements that define the behavior of a business process. They consist of Events (circles, representing something that happens), Activities (rounded rectangles, representing work that is done), and Gateways (diamonds, representing decisions or points of divergence/convergence).

- **Connecting Objects:** These objects connect the Flow Objects. They include Sequence Flows (solid arrows, showing the order of activities), Message Flows (dashed arrows, showing messages between separate process pools), and Associations (dotted lines, linking artifacts to flow objects).

- **Swimlanes:** These are used to organize activities into different groups. A Pool represents a major participant in a process (e.g., a company), while Lanes within a pool represent specific roles or departments (e.g., "Sales," "Logistics").

- **Artifacts:** These provide additional information about the process. They include Data Objects (showing data required for an activity) and Annotations (allowing for textual comments).

1.4 *Large Language Models*: Recent advances in generative AI—particularly large language models (LLMs) such as GPT-4 and Gemini have opened new possibilities for capturing procedural knowledge through natural-language interaction [5, 6]. LLMs can converse, ask probing questions, and generate structured text on demand. This makes them promising foundations for conversational assistants that elicit tacit knowledge from domain experts and translate it into formal artefacts such as BPMN diagrams.

1.5 *Generative AI applications in BPM*: The last two years have witnessed a surge of LLM-driven BPM solutions. Flowable's 2025.1 release integrates LLM-powered "AI agents" directly into executable workflows, signalling growing industrial interest [7]. Academic prototypes now automate value analysis of process models [8], articulate research agendas for LLMs across the BPM lifecycle [9], and even generate BPMN diagrams from plain-language descriptions [10]. Collectively, these works show that generative AI can lower the barrier to rigorous process documentation and analysis—an especially attractive prospect for resource-constrained SMEs.

1.6 *Research Gap and Contributions:* Existing tools either analyse already-modelled processes [8] or perform a one-shot translation of a textual narrative into BPMN [10]. What is missing is an interactive, conversational system that guides non-technical SME staff through iteratively describing their workflows and incrementally builds an accurate BPMN model. To address this gap, we present an LLM-powered chatbot that:

- Implements an interview-style methodology in which the chatbot incrementally questions the user, clarifies ambiguities, and structures the resulting knowledge.
- Uses Gemini 2.5 Pro, one of the leading state of the art LLMs as the reasoning engine [6] and a lightweight web interface to render BPMN in real time; prompt-engineering techniques ensure the LLM's output conforms to the BPMN 2.0 schema.
- Demonstrates the chatbot in an equipment-maintenance scenario, showing how it converts an informal narrative into a correct "as-is" model, analyses that model, and proposes an improved "to-be" variant.

*Paper outline*: Section 2 details the chatbot architecture. Section 3 presents the evaluation case study and results. Section 4 discusses findings, limitations and future enhancements. Section 5 concludes and outlines avenues for future research.

## 2 System Design

The system is built on a modular architecture that separates the conversational interface, the AI-driven intelligence layer, and the process visualization engine. This design ensures flexibility and allows the system to be maintained and upgraded efficiently. The core workflow is an iterative loop: a user describes a business process in natural language, the system generates a formal BPMN 2.0 diagram, and the user refines it through continued dialogue.

### 2.1. Implementation and Technology Stack

The prototype's architecture is realized using three key components:

- Conversational Frontend: The user interface is built with Gradio, an open-source Python framework that enables the rapid development of web-based UIs for machine learning models. We implemented a simple split-screen layout featuring a chat window and a diagram canvas, which proved ideal for prototyping and facilitating the interactive dialogue between the user and the AI.

- Client-Side Visualization: Process diagrams are rendered in the browser using bpmn-js, a powerful, open-source JavaScript toolkit for BPMN 2.0. Because all rendering and editing logic executes on the client side, users get instant visual feedback and can interact with the diagram without server lag. The final model can be exported directly from the interface as a standards-compliant bpmn or .xml file for use in automation engines, or as a high-resolution SVG image for inclusion in reports and documentation.

- Intelligence Layer: The system's core logic is powered by Google's Gemini 2.5 Pro API.

### 2.2. Prompt Engineering for Structured Output

Achieving reliable and functional output from the LLM required considerable prompt engineering. Early iterations produced inconsistent or malformed XML. The core system prompt was therefore meticulously refined over multiple iterations to compel the model into a highly structured and predictable response format.

The final prompt instructs the Gemini model to always respond with exactly two fenced code blocks:

1. A xml block containing the complete, standards-compliant BPMN 2.0 XML for the process diagram.

2. A json block containing an array of overlay objects for adding comments to the diagram.

This dual-output strategy is a key design feature. It cleanly separates the process model's structure (the XML) from any analytical commentary (the JSON). When a user requests feedback on a process, as in the evaluation scenario (Figure 3B), the LLM is instructed to populate the JSON array with its observations, each tied to a specific element ID in the BPMN diagram. The bpmn-js frontend then parses this JSON and renders the comments as visual annotations. With each subsequent turn, the user's new request is sent along with the entire current XML, instructing the model to modify the existing diagram rather than starting over. This makes the system a truly interactive and state-aware modeling partner.

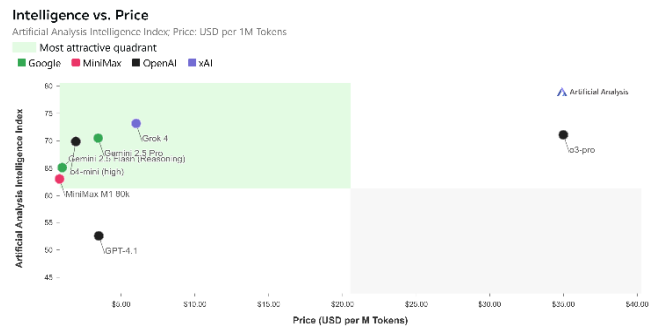

*Figure 2 The Intelligence vs. Price scatter plot from Artificial Analysis (https://artificialanalysis.ai), showing the competitive positioning of various LLMs. The y-axis represents the "Artificial Analysis Intelligence Index", and the x-axis shows the price per million tokens. The highlighted upper-left "most attractive quadrant" contains models with a favourable performance-to-cost ratio, including the Gemini 2.5 Pro used in the prototype*

### 2.3 Cost-Effectiveness and SME Accessibility

A primary design goal was to create a solution that is not only powerful but also economically viable for SMEs with limited budgets. This low-cost orientation is reflected in every layer of the technology stack. The foundation of the application is built entirely on open-source software- including the Python backend, the Gradio interface, and the bpmn-js visualization library- which carries no licensing or subscription fees.

The only direct operational cost is the pay-per-use API for the intelligence layer. The choice of Google's Gemini 2.5 Pro was a strategic decision to maximize the "intelligence-per-dollar" ratio. As shown in the Artificial Analysis benchmark (Figure 2), Gemini 2.5 Pro occupies the "most attractive quadrant" of performance versus cost, delivering state-of-the-art reasoning at an effective price of $3.44 per million tokens. This allows the system to reliably handle complex process logic without incurring prohibitive expenses. Furthermore, for less demanding tasks or for SMEs prioritizing higher throughput, the architecture supports scaling down to even more cost-effective models like Gemini 2.5 Flash, which offers a capable intelligence index at only $0.26 per million tokens.

*Figure 3 Screenshots from an interactive modelling session, detailed in Section 3.2*

By combining a zero-cost, open-source foundation with a strategically priced, high-performance AI model, this approach provides SMEs with an accessible on-ramp to sophisticated process modeling and analysis, directly aligning with the need for low-cost automation solutions.

## 3 Preliminary Prototype Evaluation

This section evaluates the prototype's performance in a real-world scenario to test its ability to capture, analyze, and enhance a complex business process.

**3.1 Evaluation Scenario:** Reactive and Untracked Equipment Maintenance

The chosen case study focuses on a common and inefficient process for managing unplanned equipment downtime, a critical issue that directly impacts production capacity. The existing workflow is almost entirely reactive and relies on informal, manual communication. The process begins when an operator verbally reports a machine fault to a supervisor. This triggers a series of inefficient steps, including the dispatch of a non-specialist technician, manual inventory checks, and paper-based purchase requests.

This scenario was selected because it involves multiple actors, hand-offs, and decision points, making it an ideal test for the LLM's ability to handle complex control-flow and data-flow details. The lack of a formalized, digitally-tracked system leads to poor visibility for production managers, who cannot predict repair times or adjust schedules effectively. Furthermore, the absence of data capture prevents valuable analysis of machine reliability and failure patterns, hindering any transition toward a more cost-effective preventive maintenance strategy.

3.2 Interactive Modeling Session

The evaluation was conducted through an interactive dialogue between a user and the chatbot, as illustrated in Figure 3. The goal was to transform the tacit knowledge of the informal maintenance process into a structured "AS-IS" model and then generate an improved "TO-BE" model.

**Step 1**: Generating the "AS-IS" Model

The user initiated the conversation by describing the entire reactive maintenance process in a single, detailed message:

> *"An assembly line operator hears a strange noise from a machine. He shuts it down and walks over to the maintenance workshop to report it. He finds the maintenance supervisor and verbally explains the problem. The supervisor grabs the nearest available technician, who may or may not be the expert on that specific machine, and sends them to investigate. The technician walks to the machine, inspects it, and determines a specific bearing has failed. They walk back to the workshop to check the physical spare parts inventory. The part is not in stock. The technician fills out a paper purchase request form and leaves it on the procurement manager's desk. They then return to the machine and leave a handwritten 'Awaiting Part' sign on it before moving to another task. There is no formal update to the operator or production manager"*

The system processed this text and generated the initial "AS-IS" BPMN diagram (Figure 3A). This first-generation task took 230 seconds.

**Step 2**: Automated Process Analysis

Next, the user prompted the model to analyse the newly created process by asking:

> *"any comments on issues you see in the current process?"*

In response, the LLM identified inefficiencies and bottlenecks in the workflow, annotating the diagram with comments (highlighted in green in Figure 3B). This analysis and comment generation took 240 seconds.

**Step 3**: Creating the "TO-BE" Model

Finally, the user requested a solution to the identified problems:

> *"Create a To-Be model which solves the issues with the current model"*

The model then generated a revised, optimized process diagram (Figure 3C) that addressed the previously highlighted issues. The generation of this "TO-BE" model was completed in 250 seconds.

## 4. Analysis & Discussion

The preliminary evaluation of the prototype demonstrates the feasibility of using an LLM-based chatbot to convert tacit operational knowledge into formal BPMN diagrams. However, the development and testing process revealed several key challenges and opportunities for future refinement. This section analyzes the choice of model, system performance, the nature of LLM-generated outputs, and promising directions for future work.

**4.1 Model Selection and Performance**

The choice of the underlying Large Language Model is critical to the system's success. The prototype was developed using Google's Gemini 2.5 Pro, which was selected for its favorable position in the "most attractive" quadrant of the intelligence-versus-price analysis, balancing high capability with cost-efficiency. This proved to be a crucial decision.

During testing, an alternative model, Gemini 2.5 Flash, was considered as a more cost-effective option. However, experiments showed that while cheaper, Gemini Flash often produced incorrectly formatted XML that failed to render. This finding suggests that for tasks requiring strict adherence to a complex schema like BPMN 2.0, the higher reasoning capability of a state-of-the-art model like Gemini 2.5 Pro is not a luxury but a necessity for reliable output. The slight cost increase is justified by the significant improvement in reliability and the avoidance of frustrating generation failures.

### 4.2 System Latency and Token Efficiency

A significant challenge observed during the evaluation was latency. For the complex maintenance scenario, the total time to generate the initial model, add comments, and create the "TO-BE" version was over 12 minutes (230s + 240s + 250s). In some cases, generation took up to 10 minutes for a single, complex output. This delay is largely attributable to the verbosity of the BPMN XML format. Generating large, structured text files is a token-intensive task that directly impacts both cost and processing time.

A potential solution, noted for future exploration, is to adopt an intermediate representation. The LLM could be prompted to first generate the process logic in a more concise format, such as JSON or even a specialized Markdown syntax. This intermediate format could then be converted into the final BPMN 2.0 XML by a separate, deterministic function.

### 4.3 Managing Non-Deterministic Outputs

A core characteristic of LLMs is that their output is inherently stochastic, not deterministic. The same input prompt can produce slightly different results on subsequent runs. The prototype currently uses a temperature setting of 1, which encourages higher creativity and variability. While this can be seen as a drawback for consistency, it can also be leveraged as a feature.

Instead of aiming for a single, consistent output, future work could embrace this stochasticity to enhance the system's utility. By running the generation for a "TO-BE" model multiple times, the system could propose several different and varied improvement scenarios. This would present the user with a range of strategic options to explore, transforming the tool from a simple modeler into a more powerful brainstorming and decision-support partner.

### 4.4 Future Directions and Enhancements

The current prototype establishes a strong foundation, but there are several promising avenues for future development:

- Agentic AI for Proactive Dialogue: The current workflow requires the user to describe the process in detail upfront. A more advanced, agentic version of the AI could build the model through a more natural, back-and-forth conversation. By using reasoning and tool-use capabilities, the agent could recognize gaps in its understanding and proactively ask clarifying questions. For example, if a user states, "The technician fixes the machine," the agent could ask, "What happens if the required part is not in stock?" or "How is the production manager notified once the machine is back online?". This would guide non-technical users through the process of building a complete model, making the knowledge capture more intuitive and less demanding.

- On-Premise Models for Data Security: For many SMEs, business processes are a form of intellectual property. The risk of IP leakage, however small, can be a barrier to adopting cloud-based AI tools. Future versions should explore the use of powerful open-source models that can be run on-premise, giving companies full control over their sensitive data.

- Agentic AI for Layout Optimization: A limitation of the current system is that the LLM only generates the BPMN code; it has no awareness of the final visual layout. Future work could explore an "agentic" approach where the AI is given the tools to "see" the rendered diagram it creates. It could then iteratively adjust the XML to improve the diagram's readability and aesthetic layout, making the output more immediately useful without manual adjustments.

- Multimodal Inputs for Frictionless Capture: To further lower the barrier to process documentation, future iterations could move beyond text-based descriptions. The system could be enhanced to accept multimodal inputs, such as analyzing video recordings of shop-floor activities to automatically map out a process. This would represent a significant leap forward in capturing the deeply ingrained, tacit knowledge of physical tasks.

- Expert Evaluation of "TO-BE" Models: While the generated "TO-BE" model in Figure 3C appears more streamlined, the current study lacks a formal evaluation of its quality. Future work should involve subjecting the AI-suggested process improvements to review by lean manufacturing experts or process engineers to validate their real-world effectiveness and ensure they align with industry best practices.

## 5. Conclusion

This paper demonstrates the viability of an AI-powered chatbot for converting tacit operational knowledge into formal BPMN 2.0 diagrams, offering SMEs a practical path to improving efficiency and preserving institutional knowledge. The prototype successfully modelled a complex "AS-IS" process, analysed its flaws, and generated an improved "TO-BE" version. Key challenges were identified, including significant system latency from verbose XML generation and the non-deterministic nature of LLM outputs. These findings highlight that while feasible, the reliability of such systems is highly dependent on the reasoning capabilities of the underlying model.

These challenges inform a clear roadmap for future development. The issue of latency may be mitigated by adopting more concise intermediate data formats, while the model's inherent variability can be leveraged as a feature to generate multiple, varied improvement scenarios for user consideration. The most promising future work lies in creating a more proactive, agentic AI that can guide users through clarifying dialogue, run on-premise for enhanced data security, and even accept multimodal inputs like video to further reduce the friction of process capture.